\documentclass[10pt,twocolumn,letterpaper]{article}

\usepackage[pagenumbers]{cvpr} 

\definecolor{cvprblue}{rgb}{0.21,0.49,0.74}
\usepackage[pagebackref,breaklinks,colorlinks,allcolors=cvprblue]{hyperref}

\def\paperID{*****} 
\def\confName{CVPR}
\def\confYear{2026}

\title{One Layer's Trash is Another Layer's Treasure: Adaptive Layer-wise Visual Token Selection in LVLMs}

\author{
    Yongru Chen$^{1}$\footnotemark[1] \quad
    Kai Zhang$^{3,1}$\footnotemark[1] \quad
    Zeliang Zong$^{1}$ \quad
    Yuchen Lu$^{2}$ \\
    Wenming Tan$^{1}$\footnotemark[2] \quad
    Ye Ren$^{1}$ \quad
    Jilin Hu$^{3}$\footnotemark[2] \\
    $^{1}$Hikvision Research Institute \quad
    $^{2}$Peking University \quad
    $^{3}$East China Normal University
}

\begin{document}
\maketitle
\footnotetext[1]{Equal contribution. † Corresponding authors.}  
\begin{abstract}
Large Vision-Language Models (LVLMs) have achieved remarkable success across diverse multimodal tasks, yet their practical deployment remains constrained by the computational burden arising from lengthy visual tokens. While visual token pruning has emerged as a promising solution, existing methods suffer from a fundamental limitation: once tokens are pruned at a specific layer, they become inaccessible to all subsequent layers, leading to premature information loss that can compromise model performance. Through empirical studies, we observe that different layers exhibit distinct visual region focus, indicating a varying optimal token subset across layers. Motivated by this insight, we propose \textbf{A}daptive \textbf{L}ayer-wise \textbf{V}isual \textbf{T}oken \textbf{S}election (ALVTS), a novel framework that breaks away from the conventional static token pruning paradigm. ALVTS incorporates a lightweight token selector to identify and route important tokens for further processing, while allowing less important tokens to skip the layer, thus minimizing computational redundancy. These two streams of tokens are seamlessly reintegrated before being fed into subsequent layers, facilitating adaptive compression across the entire model. Grounded in our importance consistency constrained low-rank approximation, the proposed token selection module closely emulates the full attention mechanism, effectively capturing its essential patterns without requiring model retraining. Extensive experiments on LLaVA-1.5, LLaVA-NeXT, and Qwen2.5-VL validate the effectiveness of our method. With an 89\% token compression ratio, ALVTS retains 96.7\% of the original model's accuracy, achieving a superior efficiency-accuracy trade-off for LVLM inference.
\end{abstract}    
\section{Introduction}
\label{sec:intro}
The remarkable success of Large Language Models (LLMs)~\cite{zhang2022opt, touvron2023llama, yang2025qwen3, cai2024internlm2} in natural language processing has catalyzed the development of Large Vision-Language Models (LVLMs)~\cite{liu2024llavanext, bai2025qwen2, wang2025internvl3, lin2024video}, which aim to bridge visual and textual modalities for a wide range of multimodal tasks, such as visual question answering~\cite{antol2015vqa, goyal2017making, marino2019ok}, image captioning~\cite{chen2024sharegpt4v, nguyen2023improving, chen2022visualgpt}, and object grounding~\cite{ma2024groma, peng2023kosmos, xu2025mc}. While LVLMs have demonstrated impressive performance, the substantial number of visual tokens poses significant computational challenges. For instance, LLaVA-NeXT~\cite{liu2024llavanext} can generate up to 2880 tokens when processing a $672 \times 672$ image, whereas textual inputs typically contain fewer than 100 tokens. This imbalance results in a disproportionate computational load dominated by the visual modality~\cite{zhang2025vscan, shao2025tokens}, which significantly limits the practical deployment of LVLMs, especially in resource-constrained environments~\cite{yao2024minicpm, sharshar2025vision}.
\begin{figure*}[t]
  \centering
  \includegraphics[width=\textwidth]{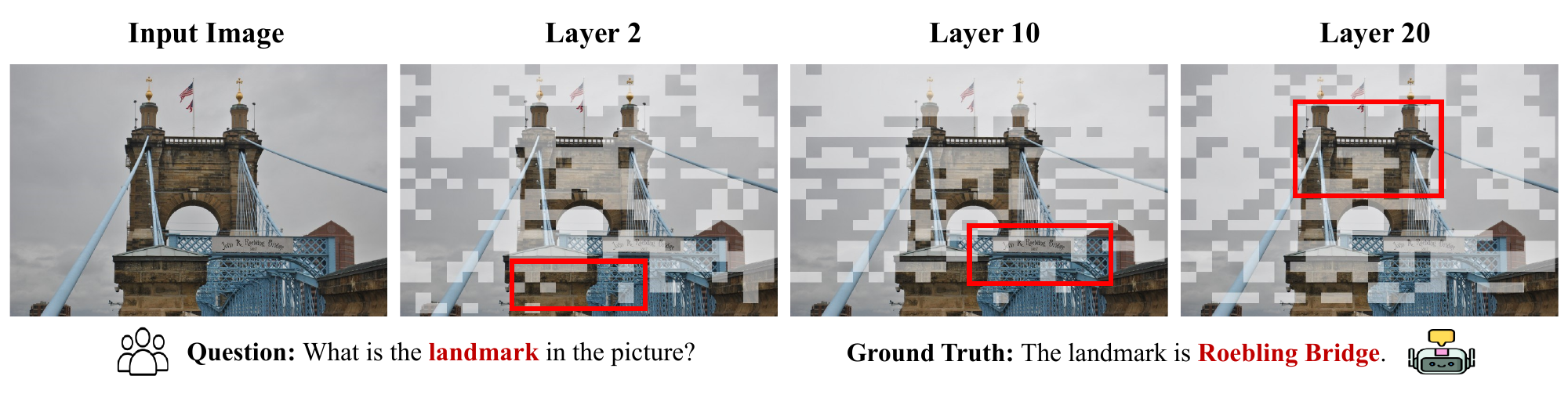}
  \caption{Layer-wise visual token attention patterns of FastV. Colored and grey regions indicate important and unimportant tokens, respectively. Red boxes highlight representative attention regions in each layer.}
  \label{fig:fig1}
\end{figure*}

Previous research~\cite{chen2024image, lin2025boosting} has demonstrated that visual tokens in VLMs exhibit significant redundancy, and visual token pruning has emerged as a promising solution to address the computational bottleneck of VLMs. Many approaches achieve this by estimating token importance from attention signals at specific modules. Common choices include the [CLS] token attention~\cite{wang2024cls, zhang2025beyond, yang2025visionzip} in the ViT backbone~\cite{dosovitskiy2020image}, or the text-to-vision attention~\cite{chen2024image, yin2025lifting, ye2025fit} in the LLM component. While intuitive, these methods share a common limitation: once tokens are pruned at a specific layer, they become inaccessible to all subsequent layers. This leads to the irrecoverable loss of visual information that may prove crucial in later layers, thereby degrading the model's generation quality.

We illustrate this issue using the pioneering FastV~\cite{chen2024image} algorithm as an example. FastV prunes tokens based on attention scores derived from the layer 2 of the LLM. Following this methodology, we visualize the image tokens that receive high attention at each layer, as shown in \cref{fig:fig1}. We observe clear cross-layer heterogeneity: layer 2 attends predominantly to the  lower bridge structure, layer 10 captures the bridge name plaque, and layer 20 concentrates on the tower structure, demonstrating distinct attention patterns across layers. Only by integrating important visual information across multiple layers can the model generate accurate responses. Conversely, pruning tokens at a single layer risks discarding instruction-relevant regions, leading to incorrect predictions.

Motivated by the above observation, we propose ALVTS, a novel and efficient framework for visual token compression. Unlike existing approaches that retain a fixed subset of compressed tokens across all layers, ALVTS performs layer-wise dynamic token selection. By allowing each layer to process only the most relevant tokens for its specific visual focus, we can dramatically reduce redundant computations without sacrificing the model's ability to capture essential visual semantics.

\begin{figure*}[t]
  \centering
  \includegraphics[width=\textwidth]{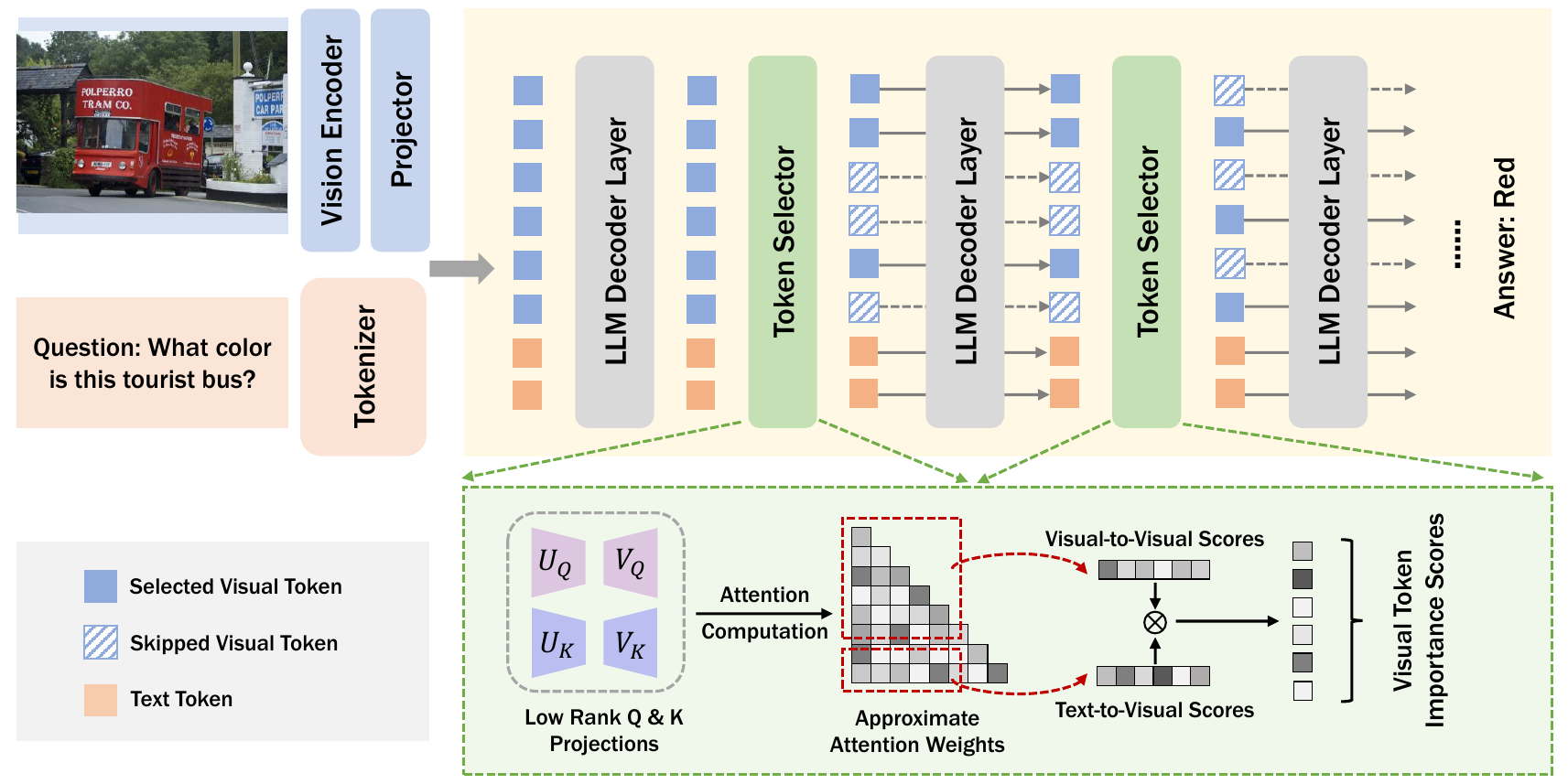}
  \caption{Overview of ALVTS. The framework performs adaptive layer-wise visual token selection. At each LLM decoder layer, a lightweight token selector computes importance scores using low-rank approximations of full attention mechanism. Importance tokens participate in the layer's computation while less important tokens skip the layer. Both streams of tokens are reintegrated before the next layer, enabling adaptive compression across the model.}
  \label{fig:algorithm}
\end{figure*}

A key challenge in layer-wise token selection is determining which tokens are most important without performing full attention computations, as this would incur excessive computational overhead~\cite{keles2023computational, ainslie2023gqa}. To address this, we design a lightweight token selector that performs dynamic token selection at each layer (see \cref{fig:algorithm}). Specifically, before visual tokens are fed into a decoder layer, they are first processed by the selector which computes importance scores for each token. Then we dynamically select a subset of high-scoring tokens to participate in the current layer's computation, while low-importance tokens skip the layer directly, avoiding unnecessary operations. At the output of the decoder layer, the processed tokens are seamlessly merged with the skipped tokens, restoring the complete token sequence for subsequent layers. This process repeats independently at each layer, enabling adaptive and context-aware token pruning throughout the model. Crucially, our token selector is theoretically grounded in importance consistency constrained low-rank approximation, which enables it to closely emulate the full attention mechanism in token selection behavior. This design eliminates the need for retraining the models, making ALVTS practical for real-world deployment.

To validate the effectiveness of ALVTS, we conduct extensive experiments on popular LVLM architectures, including LLaVA-1.5~\cite{liu2024improved}, LLaVA-NeXT~\cite{liu2024llavanext}, and Qwen2.5-VL~\cite{bai2025qwen2}. Our experimental results demonstrate that ALVTS achieves a favorable balance between efficiency and accuracy. When compressing 89\% of visual tokens, ALVTS maintains 96.7\% of the original model's performance and delivers a 1.6$\times$ inference speedup.

Our main contributions are summarized as follows:
\begin{itemize}
\item We reveal a fundamental limitation in existing token pruning strategies: once tokens are pruned at a specific layer, they become inaccessible to all subsequent layers, leading to irreversible information loss that can be critical for deeper layers.

\item We introduce ALVTS, which performs adaptive visual token selection at each layer rather than using a globally fixed token subset. This approach eliminates redundant computations while preserving essential visual semantics throughout the model.

\item We develop an efficient low-rank token selector that maintains high fidelity to the full attention mechanism without requiring model retraining.

\item Comprehensive evaluations across various LVLMs and benchmarks demonstrate that ALVTS achieves superior efficiency-accuracy trade-offs.
\end{itemize}
\section{Related Works}

\subsection{Large Vision-Language Models}
Several representative architectures have advanced the field of Large Vision-Language Models significantly. LLaVA~\cite{liu2024improved} combines CLIP vision encoders~\cite{radford2021learning} with language models through trainable projection layers, demonstrating effective vision-language alignment via instruction tuning. LLaVA-NeXT~\cite{liu2024llavanext} introduces the AnyRes technique to process high-resolution images, balancing visual detail preservation with computational efficiency. Qwen2.5-VL~\cite{bai2025qwen2} employs the naive dynamic resolution mechanism that can handle arbitrary image resolutions, which ensures consistency between the model input and the inherent information in images. The InternVL series~\cite{chen2024expanding, wang2025internvl3} divides images into a variable number of tiles according to aspect ratio, supporting up to 4K resolution input. Despite these impressive capabilities, LVLMs face substantial efficiency challenges as the number of visual tokens increases dramatically~\cite{shao2025tokens}.

\subsection{Visual Token Compression for LVLMs}
Existing methods primarily fall into attention-based and diversity-based categories. Attention-based approaches leverage attention mechanisms~\cite{vaswani2017attention} to assess token importance and guide pruning decisions. Methods utilizing vision encoder attention, such as VTC-CLS~\cite{wang2024cls}, VisPruner~\cite{zhang2025beyond}, and VisionZip~\cite{yang2025visionzip}, exploit [CLS] token attention to identify dominant tokens that aggregate substantial visual information. Another category of methods exploits attention within language models to evaluate token importance. For instance, FastV~\cite{chen2024image} prunes visual tokens with low attention scores after the filtering layer. PyramidDrop~\cite{xing2024pyramiddrop} observes that attention to visual tokens progressively concentrates across layers, and implements multi-stage token reduction with a pyramid dropping ratio. FitPrune~\cite{ye2025fit} formulates token pruning as a statistical distribution fitting problem, minimizing the divergence of both text-visual and intra-visual attention before and after pruning based on a small calibration set. Alternatively, diversity-based methods prioritize maximizing token diversity to reduce redundancy. DivPrune~\cite{alvar2025divprune} formulates pruning as a Max-Min Diversity Problem~\cite{porumbel2011simple, resende2010grasp} to maximize minimum pairwise distances among selected tokens, and DART~\cite{wen2025stop} selects tokens with the lowest similarity to pivot tokens. Nevertheless, all these methods perform static, irreversible pruning, potentially causing premature loss of tokens that become important in deeper layers.
\section{Method}
\subsection{Preliminary}
LVLMs are designed to process both text and visual inputs. For the text input, it is embedded into $M$ textual tokens $X_T = [t_1, t_2, \ldots, t_M] \in \mathbb{R}^{M \times D}$ through the language model's embedding layer. For the visual input, a vision encoder extracts visual features, which are subsequently projected into $N$ visual tokens $X_V = [v_1, v_2, \ldots, v_N] \in \mathbb{R}^{N \times D}$ via a projector layer, where typically $N \gg M$.

These text and visual tokens are concatenated as $X = [X_T, X_V] \in \mathbb{R}^{(M+N) \times D}$ and passed through the language model for autoregressive generation. At each step $t$, the model generates an output token $y_t$ based on all previously generated tokens: $y_t \sim P(\cdot \mid y_{<t}, X_T, X_V)$, where $P(\cdot)$ denotes the conditional probability distribution from the language model, and $D$ represents the hidden dimension.

\subsection{ALVTS Framework}
\label{sec:ALVTS Framework}
In this section, we present ALVTS, a novel framework that performs layer-wise dynamic token selection. At each layer of the language model, ALVTS computes importance scores for all visual tokens using a low-rank approximation~\cite{golub1987generalization} of the attention mechanism. Based on these scores, the most informative visual tokens are selected to participate in the layer's computation along with all text tokens, while the remaining visual tokens skip the layer. After processing, all tokens are reassembled to form the complete input sequence for the next layer. The pipeline is illustrated in \cref{fig:algorithm}.

\textbf{Token Importance Estimation via Low-Rank Approximation.}
We employ a token selector $\mathcal{P}$ to compute importance scores for each visual token. To avoid the overhead of full attention computation, we design the selector using low-rank approximations of the layer's query and key projection matrices in the self-attention module:
\begin{equation}
\tilde{W}_Q = U_Q V_Q, \quad \tilde{W}_K = U_K V_K,
\end{equation}
where $U_Q, U_K \in \mathbb{R}^{D \times R}$ and $V_Q, V_K \in \mathbb{R}^{R \times D}$, with $R \ll D$ being the rank. Given the input sequence $X = [X_T, X_V]$, we compute the approximate queries and keys:
\begin{equation}
\tilde{Q} = X \tilde{W}_Q^\top, \quad
\tilde{K} = X \tilde{W}_K^\top,
\end{equation}
and derive the attention weights between all tokens:
\begin{equation}
\tilde{A} = \text{softmax}\left(\frac{\tilde{Q} \tilde{K}^\top}{\sqrt{d_k}}\right) \in \mathbb{R}^{(M+N) \times (M+N)}.
\end{equation}

We partition the attention matrix $\tilde{A}$ into two components: $\tilde{A}_{V2V} \in \mathbb{R}^{N \times N}$ representing visual-to-visual attention, and $\tilde{A}_{T2V} \in \mathbb{R}^{M \times N}$ representing text-to-visual attention. For the $i$-th visual token, we compute its visual-to-visual score as the average attention it receives from all visual tokens, and its text-to-visual score as the average attention it receives from all text tokens:
\begin{align}
\mathcal{S}_{V2V}(i) &= \frac{1}{N} \sum_{j=1}^{N} \tilde{A}_{V2V}(j, i), \\
\mathcal{S}_{T2V}(i) &= \frac{1}{M} \sum_{j=1}^{M} \tilde{A}_{T2V}(j, i).
\end{align}

The final importance score is computed as the product of these two components:
\begin{equation}
\mathcal{S}(i) = \mathcal{S}_{V2V}(i) \cdot \mathcal{S}_{T2V}(i), \quad i = 1, 2, \ldots, N.
\end{equation}

This multiplicative combination ensures that tokens are deemed important only when they are relevant to both the visual and textual contexts. Tokens with low scores in either dimension are considered less critical and can be safely skipped for the current layer.

\textbf{Layer-wise Token Selection and Processing.}
Based on the computed importance scores, we perform top-$k$ selection to identify the most informative visual tokens:
\begin{equation}
X_{V}^{(\text{select})} = \text{TopK}(X_V, \mathcal{S}, k),
\end{equation}
where $X_{V}^{(\text{select})}$ contains $k$ visual tokens and $k = \lfloor r \cdot N \rfloor$ with $r$ being the selection ratio. The complementary set of skipped tokens is denoted as $X_{V}^{(\text{skip})}$, which contains the remaining $(N - k)$ visual tokens.

The selected tokens $X_{V}^{(\text{select})}$ together with all text tokens participate in the layer's computation, while $X_{V}^{(\text{skip})}$ skip the layer directly. At the output of layer $\ell$, the processed tokens and skipped tokens are recombined to form the input for the next layer. Note that we maintain the original token ordering throughout this process to preserve positional information.

A key characteristic of ALVTS is that the token selection process is performed independently at each layer. Different layers may select different subsets of visual tokens based on their specific attention patterns, which enables adaptive and context-aware token compression throughout the model.

\subsection{Token Selector Optimization}
\begin{table*}[t]
    \caption{Performance comparison of LLaVA-1.5-7B under different token reduction ratios. ``Avg.'' refers to the average performance across 8 benchmarks. Best results are shown in \textbf{bold}.}
    \label{tab:comparison_llava_1.5_7b}
    
    \centering
    \begin{tabular}{l|cccccccc|c}
    \toprule
    \textbf{Method} & \textbf{AI2D} & \textbf{POPE} & \textbf{VQA}$^{\text{Text}}$ & \textbf{OKVQA} & \textbf{VizWiz} & \textbf{COCO} & \textbf{NoCaps} & \textbf{RealWorld} & \textbf{Avg.} \\
    \midrule
    \multicolumn{10}{c}{\textit{Upper Bound, 576 Tokens (100\%)}} \\
    \midrule
    LLaVA-1.5-7B & 55.21 & 85.84 & 58.18 & 53.42 & 54.32 & 110.52 & 105.61 & 56.08 & 100.00\% \\
    \midrule
    \multicolumn{10}{c}{\textit{Retain 192 Tokens ($\downarrow$ 67\%)}} \\
    \midrule
    FastV {\footnotesize\textit{(ECCV-24)}} & 54.73 & 77.79 & 57.91 & 51.77 & 54.95 & 107.49 & 102.26 & 53.33 & 97.07\% \\
    VTW {\footnotesize\textit{(AAAI-25)}} & 54.99 & 74.86 & 50.60 & 30.60 & 50.66 & 39.85 & 26.66 & 50.46 & 71.95\% \\
    PDrop {\footnotesize\textit{(CVPR-25)}} & 54.40 & 84.67 & 57.11 & \textbf{52.37} & 54.42 & 109.28 & 103.76 & 53.20 & 98.19\% \\
    DART {\footnotesize\textit{(EMNLP-25)}} & 54.73 & 82.45 & 57.40 & 51.50 & 54.95 & 108.93 & 103.84 & 54.25 & 98.13\% \\
    \textbf{ALVTS (\textit{Ours})} & \textbf{55.44} & \textbf{85.72} & \textbf{58.29} & 52.12 & \textbf{55.05} & \textbf{111.68} & \textbf{104.50} & \textbf{54.64} & \textbf{99.60\%} \\
    \midrule
    \multicolumn{10}{c}{\textit{Retain 128 Tokens ($\downarrow$ 78\%)}} \\
    \midrule
    FastV {\footnotesize\textit{(ECCV-24)}} & 54.02 & 72.10 & 56.95 & 49.90 & 55.14 & 101.99 & 97.06 & 51.11 & 93.75\% \\
    VTW {\footnotesize\textit{(AAAI-25)}} & 53.04 & 41.25 & 44.26 & 22.87 & 50.76 & 9.50 & 5.76 & 44.05 & 56.13\% \\
    PDrop {\footnotesize\textit{(CVPR-25)}} & 53.92 & 81.82 & 56.18 & 50.27 & 52.35 & 107.86 & 97.01 & 51.37 & 95.13\% \\
    DART {\footnotesize\textit{(EMNLP-25)}} & 54.37 & 80.26 & 56.64 & 51.17 & \textbf{55.28} & 107.80 & \textbf{102.86} & \textbf{53.99} & 97.26\% \\
    \textbf{ALVTS (\textit{Ours})} & \textbf{54.57} & \textbf{85.99} & \textbf{57.92} & \textbf{52.13} & 54.90 & \textbf{110.32} & 102.30 & \textbf{53.99} & \textbf{98.77\%} \\
    \midrule
    \multicolumn{10}{c}{\textit{Retain 64 Tokens ($\downarrow$ 89\%)}} \\
    \midrule
    FastV {\footnotesize\textit{(ECCV-24)}} & 53.56 & 59.55 & 55.32 & 45.14 & 54.91 & 82.62 & 79.01 & 47.32 & 85.13\% \\
    VTW {\footnotesize\textit{(AAAI-25)}} & 50.26 & 31.67 & 42.85 & 19.96 & 50.34 & 9.80 & 6.06 & 43.53 & 52.98\% \\
    PDrop {\footnotesize\textit{(CVPR-25)}} & 53.30 & 72.39 & 53.36 & 45.10 & 51.56 & 99.26 & 85.45 & 47.97 & 88.52\% \\
    DART {\footnotesize\textit{(EMNLP-25)}} & 53.59 & 74.02 & 54.53 & 47.99 & 55.46 & 103.89 & \textbf{97.70} & 50.85 & 93.27\% \\
    \textbf{ALVTS (\textit{Ours})} & \textbf{54.63} & \textbf{83.87} & \textbf{57.41} & \textbf{49.66} & \textbf{55.65} & \textbf{105.89} & 97.60 & \textbf{53.20} & \textbf{96.73\%} \\
    \bottomrule
    \end{tabular}
\end{table*}

First, we initialize the low-rank projection matrices $U_Q, V_Q, U_K, V_K$ using Singular Value Decomposition (SVD) of the original projection matrices~\cite{demmel1997applied}. Specifically, for the query projection $W_Q \in \mathbb{R}^{D \times D}$, we perform SVD to obtain $W_Q = U_{\text{full}} \Sigma V_{\text{full}}^{\top}$, where $\Sigma$ contains the singular values in descending order. We retain the top $R$ singular value components to construct the initial low-rank matrices $U_Q$ and $V_Q$:
\begin{align}
\Sigma_R^{1/2} &= \text{diag}(\sqrt{\sigma_1}, \sqrt{\sigma_2}, \ldots, \sqrt{\sigma_R}) \in \mathbb{R}^{R \times R}, \\
U_Q &= U_{\text{full}}[:, :R] \cdot \Sigma_R^{1/2} \in \mathbb{R}^{D \times R}, \\
V_Q &= \Sigma_R^{1/2} \cdot V_{\text{full}}[:R, :]^{\top} \in \mathbb{R}^{R \times D}.
\end{align}

The same procedure is applied to initialize $U_K$ and $V_K$ from the original key projection $W_K$.

After initialization, we further adapt the selector to align its importance scoring behavior with that of the original model. Specifically, for each input sample, we derive the reference token importance scores $\mathcal{S}^*$ from the original model's attention patterns and align our low-rank approximation $\mathcal{S}$ by minimizing the 
reconstruction error:
\begin{equation}
\mathcal{L} = \frac{1}{N} \sum_{i=1}^{N} (\mathcal{S}(i) - \mathcal{S}^*(i))^2.
\label{eq:mse_loss}
\end{equation}

Each layer's selector is optimized independently, and we only update the parameters of the selectors, while keeping the original LVLM frozen. This design is highly efficient. Taking LLaVA-1.5-7B as an example, the entire optimization process can be completed within 15 minutes. Furthermore, our token selector introduces negligible parameter overhead to the base model. The detailed rank settings, parameter counts, and overhead for different models are presented in \cref{tab:parameter_overhead}.
\section{Experiments}
\subsection{Experimental Setup}
\noindent\textbf{Models.} We evaluate ALVTS on three representative LVLMs: LLaVA-1.5 (7B and 13B variants)~\cite{liu2024improved}, LLaVA-NeXT-7B~\cite{liu2024llavanext}, and Qwen2.5-VL-3B-Instruct~\cite{bai2025qwen2}. LLaVA-1.5 employs a fixed-length visual token sequence of 576 tokens. LLaVA-NeXT-7B supports dynamic image resolutions with variable-length visual tokens, enabling more flexible visual processing. Qwen2.5-VL-3B-Instruct supports naive dynamic resolution, generating variable-length token sequences without resizing images to a fixed input size. These models provide comprehensive evaluation scenarios for our compression method.

\begin{table}[t]
    \caption{Performance comparison of LLaVA-1.5-13B. We report the average performance across 8 benchmarks, with detailed per-benchmark results provided in the Appendix.}
    \label{tab:comparison_13b}
    
    \setlength{\tabcolsep}{3pt}
    
    \begin{tabular*}{\columnwidth}{@{\extracolsep{\fill}}l|cccccc@{}}
    \toprule
    \textbf{Method} & \textbf{Ratio} & \textbf{Avg.} & \textbf{Ratio} & \textbf{Avg.} & \textbf{Ratio} & \textbf{Avg.} \\
    \midrule
    FastV & & 98.71\% & & 96.41\% & & 91.20\% \\
    VTW & & 85.19\% & & 69.16\% & & 48.35\% \\
    PDrop & 67\% & 97.88\% & 78\% & 96.56\% & 89\% & 92.66\% \\
    DART & & 99.16\% & & 97.41\% & & 93.98\% \\
    ALVTS & & \textbf{99.86\%} & & \textbf{99.12\%} & & \textbf{96.63\%} \\
    \bottomrule
    \end{tabular*}
\end{table}
\begin{table*}[t]
    \caption{Performance Comparison of LLaVA-NeXT-7B. The vanilla number of visual tokens varies by dataset due to dynamic image processing (up to 2880 tokens). ``Avg.'' refers to the average performance across 8 benchmarks. Best results are shown in \textbf{bold}.}
    \label{tab:comparison_llava_next_7b}
    \centering
    \begin{tabular}{l|cccccccc|c}
    \toprule
    \textbf{Method} & \textbf{AI2D} & \textbf{POPE} & \textbf{VQA}$^{\text{Text}}$ & \textbf{OKVQA} & \textbf{VizWiz} & \textbf{COCO} & \textbf{NoCaps} & \textbf{RealWorld} & \textbf{Avg.} \\
    \midrule
    \multicolumn{10}{c}{\textit{Upper Bound, 2880 Tokens (100\%)}} \\
    \midrule
    LLaVA-NeXT-7B & 65.32 &	86.41 &	61.39 &	44.23 & 60.62 & 99.83 & 88.28 & 57.91 & 100.0\% \\
    \midrule
    \multicolumn{10}{c}{\textit{Retain 320 Tokens ($\downarrow$ 89\%)}} \\
    \midrule
    FastV {\footnotesize\textit{(ECCV-24)}} & 62.79 & 74.47 & 57.53 & 38.91 & 57.60 & 82.56 & 74.54 & 52.94 & 89.70\% \\
    VTW {\footnotesize\textit{(AAAI-25)}} & 60.75 & 0.27 & 37.88 & 9.61 & 52.10 & 5.97 & 3.11 & 42.22 & 43.14\% \\
    PDrop {\footnotesize\textit{(CVPR-25)}} & 62.60 & 81.80 & 56.93 & \textbf{43.27} & 58.60 & \textbf{94.68} & 79.53 & 54.25 & 94.54\% \\
    DART {\footnotesize\textit{(EMNLP-25)}} & 64.48 & 81.58 & 57.02 & 40.37 & 58.79 & 91.02 & 78.89 & 55.56 & 93.84\% \\
    \textbf{ALVTS (\textit{Ours})} & \textbf{64.51} & \textbf{87.09} & \textbf{58.37} & 41.70 & \textbf{59.36} & 94.37 & \textbf{80.24} & \textbf{56.60} & \textbf{96.25\%} \\
    \bottomrule
    \end{tabular}
\end{table*}

\noindent\textbf{Benchmarks.} We conduct experiments on 8 multimodal benchmarks: AI2D~\cite{kembhavi2016diagram}, POPE~\cite{li2023evaluating}, TextVQA~\cite{singh2019towards}, OKVQA~\cite{marino2019ok}, VizWiz~\cite{gurari2018vizwiz}, COCO Caption~\cite{lin2014microsoft}, NoCaps~\cite{agrawal2019nocaps}, and RealWorldQA. These benchmarks cover diverse tasks including general VQA, text-based VQA, image captioning, and multimodal reasoning. Further details are presented in the Appendix.

\noindent\textbf{Comparison Methods.} We compare our method with 4 popular visual token compression methods. FastV~\cite{chen2024image} prunes visual tokens with low attention scores after the second layer of the LLM. VTW~\cite{lin2025boosting} strategically withdraws vision tokens at a certain layer, enabling only text tokens to engage in the computation of subsequent layers. PyramidDrop~\cite{xing2024pyramiddrop} progressively drops visual tokens across stages as token redundancy increases in deeper layers. DART~\cite{wen2025stop} prunes tokens based on their duplication with pivot tokens, which guarantees maximal information diversity. To ensure a fair comparison, we adjust the hyperparameters of each method to match the FLOPs of FastV, which performs visual token pruning once after the second layer of the LLM.

\noindent\textbf{Implementation Details.} To optimize the token selectors, we randomly select 256 samples from the LLaVA-655k dataset~\cite{liu2023visual}. To keep parameter overhead below 2\% per decoder layer, we set the rank of low-rank weight matrices to 256 for LLaVA-1.5-7B and LLaVA-NeXT-7B, and to 128 for Qwen2.5-VL-3B-Instruct (see \cref{tab:parameter_overhead}). For performance evaluation on test benchmarks, we employ the LMMs-Eval evaluation framework~\cite{zhang2025lmms} across all the baselines and models.
\begin{table}[t]
    \caption{Token selector rank settings, parameter counts, and parameter overhead per decoder layer for different models.}
    \label{tab:parameter_overhead}

    \setlength{\tabcolsep}{4pt}

    \begin{tabular*}{\columnwidth}{@{\extracolsep{\fill}}lccc@{}}
    \toprule
    \small\textbf{Model} & \small\textbf{Rank} & \small\textbf{Params/Layer} & \small\textbf{Overhead/Layer} \\
    \midrule
    LLaVA-1.5-7B & 256 & 4.2M & 2.0\% \\
    LLaVA-NeXT-7B & 256 & 4.2M & 2.0\% \\
    LLaVA-1.5-13B & 256 & 5.2M & 1.7\% \\
    Qwen2.5-VL-3B & 128 & 0.8M & 1.1\% \\
    \bottomrule
    \end{tabular*}
\end{table}

\subsection{Main Results}
\cref{tab:comparison_llava_1.5_7b} presents comprehensive evaluation results of ALVTS against existing approaches across 8 benchmarks under three different token reduction ratios. At a 67\% reduction ratio, ALVTS achieves an average performance of 99.60\%, surpassing all baseline methods, with a 1.41\% margin over the second-best method, PyramidDrop. At a more aggressive 89\% reduction ratio, our method maintains a 96.73\% average performance, consistently outperforming all baselines by 3.46\% over DART, 8.21\% over PyramidDrop, and 11.60\% over FastV. The maintained high performance on {VQA}$^{\text{Text}}$ (57.41) and VizWiz (55.65) at this extreme compression setting suggests that our dynamic token selection strategy effectively preserves text-relevant visual features and adapts well to real-world, challenging images.

To validate the scalability of our approach, we further conduct experiments on the larger LLaVA-1.5-13B model. As shown in \cref{tab:comparison_13b}, ALVTS consistently achieves superior performance across all compression ratios. At a 67\% reduction ratio, our method reaches an average performance of 99.86\%. When the compression ratio increases to 78\% and 89\%, ALVTS maintains 99.12\% and 96.63\% average performance respectively, outperforming the second best method by 1.71\% and 2.65\% at these settings. Detailed comparison results on LLaVA-1.5-13B across 8 benchmarks are presented in the Appendix.

\subsection{ALVTS with Higher Resolution}
\begin{table}[t]
    \caption{Performance Comparison on Qwen2.5-VL-3B.}
    \label{tab:comparison_qwen}

    \small
    \setlength{\tabcolsep}{3pt} 

    \begin{tabular*}{\columnwidth}{@{\extracolsep{\fill}} l|ccccc|c}
    \toprule
    \textbf{Method} & \textbf{AI2D} & \textbf{POPE} & \textbf{COCO} & \textbf{NoCaps} & \textbf{RealWorld} & \textbf{Avg.} \\
    \midrule
    \multicolumn{7}{c}{\textit{Upper Bound, All Tokens (100\%)}} \\
    \midrule
    Vanilla & 78.72 & 86.91 & 99.78 & 106.46 &	59.35 & 100.0\% \\
    \midrule
    \multicolumn{7}{c}{\textit{Token Reduction ($\downarrow$ 89\%)}} \\
    \midrule
    FastV & 68.26 &	73.98 &	72.18 &	77.78 &	49.80 & 80.23\% \\
    PDrop & 67.03 &	67.84 &	73.54 &	84.81 &	44.84 & 78.43\% \\
    ALVTS & \textbf{68.39} & \textbf{86.58} & \textbf{78.23} & \textbf{85.72} & \textbf{51.37} & \textbf{86.39\%} \\
    \bottomrule
    \end{tabular*}
\end{table}

We also present evaluations on LLaVA-NeXT-7B, a high-resolution LVLM that processes significantly more visual tokens. As shown in \cref{tab:comparison_llava_next_7b}, when retaining only 320 tokens (from up to 2880 input tokens), ALVTS preserves 96.25\% of the original model’s average performance, outperforming PyramidDrop and DART by 1.7\% and 2.4\% respectively. The advantage is particularly pronounced on the POPE benchmark, where ALVTS achieves 87.09, even slightly surpassing the vanilla model’s performance. These results demonstrate the effectiveness of ALVTS for processing high-resolution visual input.

\subsection{ALVTS with Qwen Architecture}
To verify the generalization of ALVTS beyond LLaVA-based models, we conduct experiments on Qwen2.5-VL-3B, with results shown in \cref{tab:comparison_qwen}. At an 89\% token reduction rate, ALVTS retains 86.39\% of the original performance on average, outperforming FastV (80.23\%) and PyramidDrop (78.43\%) by 6.16 and 7.96 percentage points, respectively. These results confirm that ALVTS effectively generalizes across different vision-language model architectures.
\begin{figure}[t]
  \centering
  \includegraphics[width=\linewidth]{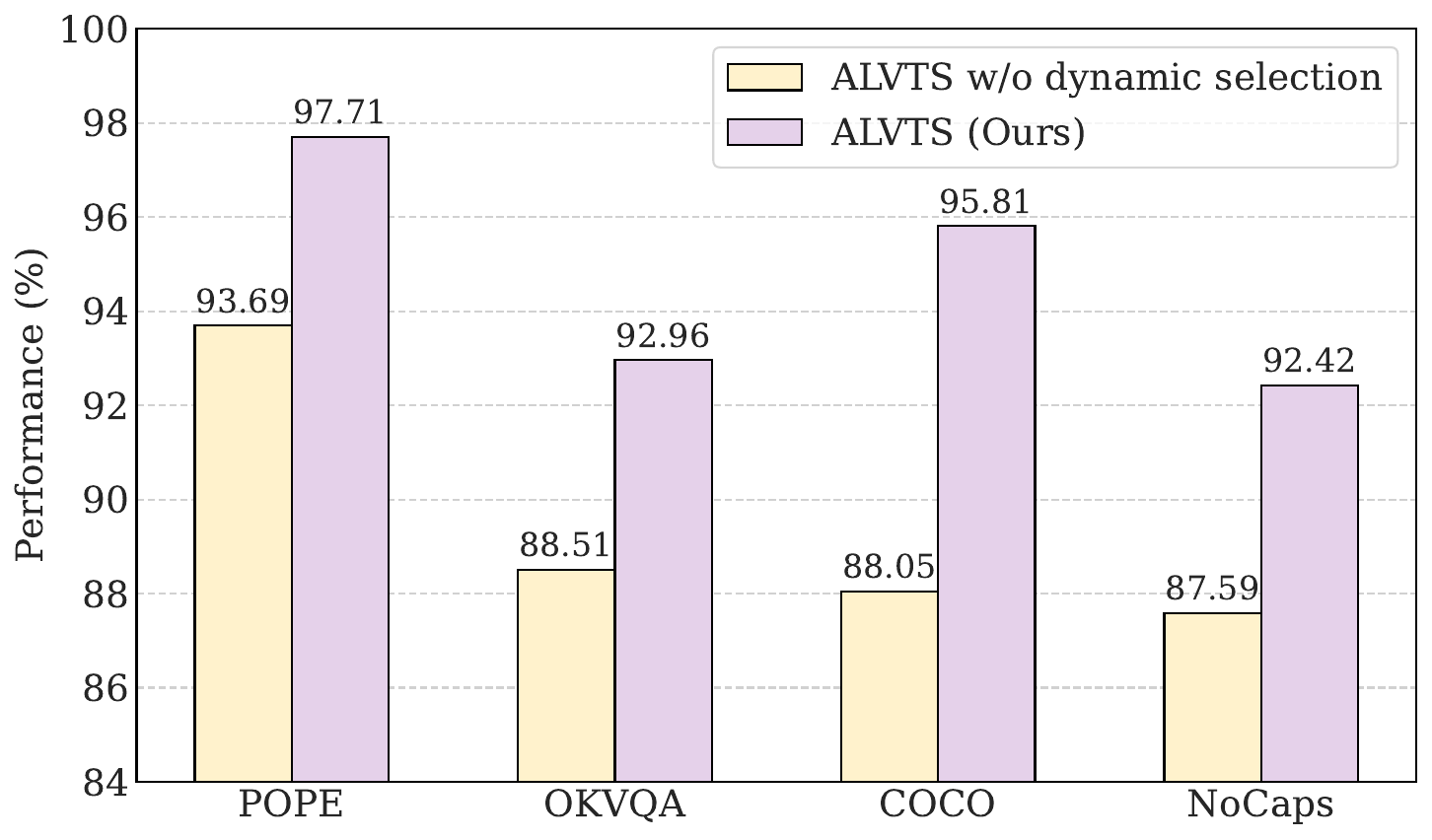}
  \caption{Ablation study of dynamic token selection mechanism.}
  \label{fig:ablation_static_dynamic}
\end{figure}

\subsection{Efficiency Analysis}
To evaluate the efficiency of ALVTS, we compare total inference time, end-to-end latency, prefill time, and accuracy on LLaVA-1.5-7B. As shown in \cref{tab:efficiency_1.5_7b}, under an 89\% token reduction ratio, ALVTS maintains a 97.7\% accuracy. Compared to the vanilla model, ALVTS decreases the average per-sample latency from 211ms to 156ms (1.35$\times$ speedup), and reduces the prefill time from 165ms to 103ms (1.6$\times$ speedup). When compared to FastV with a 60\% token reduction ratio, ALVTS delivers comparable inference speed while achieving 4.5 percentage points higher accuracy (97.7\% vs. 93.2\%). These results demonstrate that ALVTS attains a better trade-off between inference efficiency and model performance.
\begin{table}[t]
    \caption{Efficiency comparison on POPE dataset. Experiments are on LLaVA-1.5-7B with a single 4090 GPU.}
    \label{tab:efficiency_1.5_7b}

    \setlength{\tabcolsep}{4pt}

    \begin{tabular*}{\columnwidth}{@{\extracolsep{\fill}}lcccc@{}}
    \toprule
    \textbf{Method} & \textbf{Time} $\downarrow$ & \textbf{Latency} $\downarrow$ & \textbf{Prefill} $\downarrow$ & \textbf{Acc.} $\uparrow$ \\
    \midrule
    LLaVA-1.5-7B & 31:44 & 211ms & 165ms & 100\% \\
    FastV ($\downarrow$ 60\%) & 26:51 & 179ms & 126ms & 93.2\% \\
    ALVTS ($\downarrow$ 89\%) & \textbf{23:28} & \textbf{156ms} & \textbf{103ms} & \textbf{97.7\%} \\
    \bottomrule
    \end{tabular*}
\end{table}

\begin{figure}[t]
  \centering
  \includegraphics[width=\linewidth]{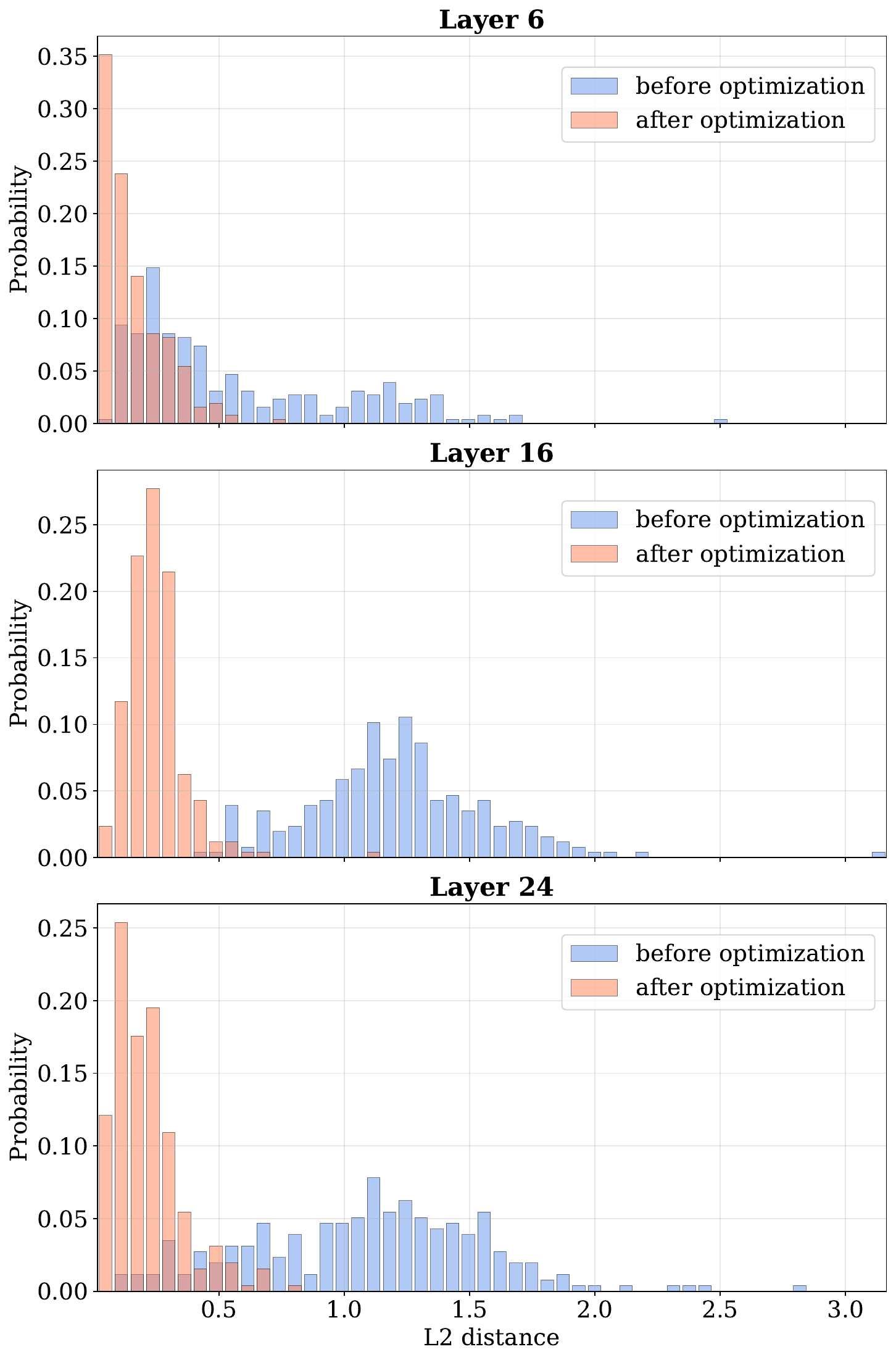}
  \caption{The distribution of $L_2$ distance between the approximate importance scores and the oracle importance scores.}
  \label{fig:histogram_plots}
\end{figure}

\subsection{Ablation Studies}
\begin{figure*}[t]
  \centering
  \includegraphics[width=\textwidth]{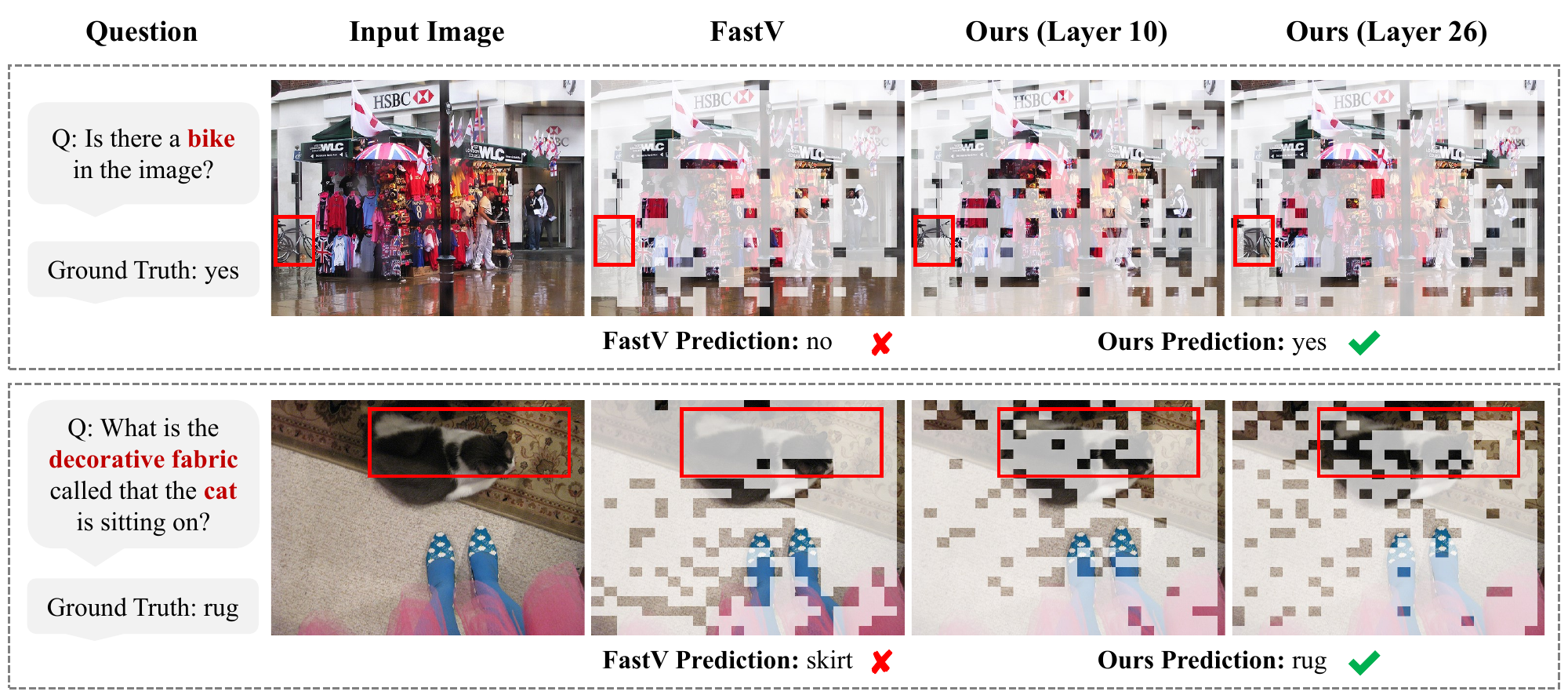}
  \caption{The case comparison between FastV and ALVTS. It presents original images alongside their pruned versions. Grey mask areas indicate discarded tokens. Red bounding boxes highlight key objects referenced in prompts.}
  \label{fig:token_selection_visualize}
\end{figure*}

\noindent\textbf{Effectiveness of Layer-wise Dynamic Token Selection.} To validate the effectiveness of the dynamic token selection mechanism, we conduct an ablation study by comparing ALVTS with a static pruning baseline (ALVTS w/o dynamic selection) on LLaVA-1.5-7B with 64 visual tokens retained. The baseline performs one-time token pruning after the second layer and maintains this pruned token set throughout all subsequent layers, following the same strategy as FastV. As illustrated in \cref{fig:ablation_static_dynamic}, the full ALVTS method consistently outperforms the static pruning baseline. The dynamic selection mechanism brings notable improvements, especially on COCO and NoCaps, achieving gains of 7.76\% and 4.83\%, respectively. These results confirm that different decoder layers exhibit varying optimal token subsets, and static pruning leads to premature information loss that compromises model performance. By allowing less important visual tokens to skip certain layers while remaining accessible for subsequent processing, ALVTS effectively balances efficiency and performance, thereby validating the necessity of the dynamic selection mechanism.

\noindent\textbf{Effectiveness of Token Selector Optimization.} To validate the effectiveness of our low-rank token selector design, we evaluate how well it approximates the importance scoring behavior of the full attention mechanism. We randomly sample 1,000 instances from the AI2D dataset and measure the $L_2$ distance between the approximate importance scores (derived from our low-rank selector) and the oracle importance scores (computed via full-rank query and key projections) of visual tokens. \cref{fig:histogram_plots} presents the $L_2$ distance distributions at several layers before and after optimization. The blue histograms depict the distance distribution obtained using SVD-initialized weights, while the orange histograms show the results after optimization with the objective defined in Eq. \eqref{eq:mse_loss}. The results reveal a pronounced improvement in approximation fidelity: the $L_2$ distance distributions become significantly more concentrated near zero after optimization, with the peak probability density shifting closer to the origin. This demonstrates that our low-rank token selector accurately captures the importance scoring patterns inherent in the full attention mechanism, thereby ensuring reliable token selection.

\subsection{Visualizations Results}
To provide a deeper understanding of our ALVTS framework, we visualize the selected visual tokens in \cref{fig:token_selection_visualize}. We present two examples where our method yields correct predictions while FastV fails. The figure illustrates the token selection results from two ALVTS layers alongside FastV’s static pruning result. For FastV, the pruning layer fails to attend to the visual regions mentioned in the prompts (e.g., ``bike'' and ``decorative fabric''), thus leading to incorrect answers. In contrast, ALVTS exhibits distinct visual region preferences across different layers. Even when shallow layers do not initially focus on the target objects, deeper layers are able to attend to and process these critical regions, ultimately ensuring accurate predictions. For instance, as shown in the bottom row, while FastV discards the tokens corresponding to the ``decorative fabric,'' ALVTS correctly identifies these critical regions at layer 10 and layer 26. This highlights the key advantage of ALVTS: by preserving token accessibility across layers and enabling dynamic token selection based on each layer's attention pattern, our method maintains robust visual grounding capabilities throughout inference. 
\section{Conclusion}
\label{sec:conclusion}
In this paper, we propose ALVTS, a novel framework that challenges the conventional paradigm of static token pruning in LVLMs. Unlike existing methods that permanently discard tokens at a specific layer, ALVTS performs adaptive layer-wise visual token selection. Each layer can process the tokens it deems important while keeping all tokens accessible throughout the model. To achieve this, we design a token selector based on importance consistency constrained low-rank approximation, which faithfully emulates the full attention mechanism in token selection behavior without model retraining. Comprehensive experiments demonstrate that ALVTS achieves superior efficiency-accuracy trade-offs, making it a practical solution for deploying LVLMs in real-world settings. By respecting the heterogeneous visual information needs across layers, we can indeed turn one layer's trash into another layer's treasure.
{
    \small
    \bibliographystyle{ieeenat_fullname}
    \bibliography{main}
}


\end{document}